# Specialized Deep Residual Policy Safe Reinforcement Learning-Based Controller for Complex and Continuous State-Action Spaces

Ammar N. Abbas[1,2], Georgios C. Chasparis[1], and John D. Kelleher[2,3]

*Abstract*— Traditional controllers have limitations as they rely on prior knowledge about the physics of the problem, require modeling of dynamics, and struggle to adapt to abnormal situations. Deep reinforcement learning has the potential to address these problems by learning optimal control policies through exploration in an environment. For safety-critical environments, it is impractical to explore randomly, and replacing conventional controllers with black-box models is also undesirable. Also, it is expensive in continuous state and action spaces, unless the search space is constrained. To address these challenges we propose a specialized deep residual policy safe reinforcement learning with a cycle of learning approach adapted for complex and continuous state-action spaces. Residual policy learning allows learning a hybrid control architecture where the reinforcement learning agent acts in synchronous collaboration with the conventional controller. The cycle of learning initiates the policy through the expert trajectory and guides the exploration around it. Further, the specialization through the input-output hidden Markov model helps to optimize policy that lies within the region of interest (such as abnormality), where the reinforcement learning agent is required and is activated. The proposed solution is validated on the Tennessee Eastman process control.

## I. INTRODUCTION

The principal objectives of a control system are to either (i) track a setpoint as accurately as possible or (ii) respond to external disturbances and restore normal operation as efficiently as possible. The process industry primarily uses conventional controllers such as Model Predictive Control (MPC) or Proportional Integral Derivative (PID) controllers. These controllers require physical/mathematical modeling of the system and are applied to isolated systems, avoiding issues related to complex processes (with unmodeled dynamics) or multi-component systems. These controllers also require continuous maintenance and tuning, and their performance may decrease over time due to system drift or set-point changes [1]–[3].

Deep Reinforcement Learning (DRL)—RL combined with Deep Neural Networks (DNN)—has been successfully applied in prior work to control continuous state-action spaces [4]. DRL learns by trial and error through interacting with the environment to learn the optimal strategy in a model-free setting [5], [6]. DRL has previously been considered in the field of process control to overcome the problems encountered by conventional controllers and apply it to non-linear stochastic control problems [7], [8]. DRL can be used to train a real-time, model-free, self-learning, and adaptive controller which is an ideal combination for complex process industries. In particular, DRL-based controllers can avoid the need for hand-crafted feature descriptors, controller tuning, deriving control laws, and developing complex mathematical models.

One challenge with using DRL for process control is that most previous research on the topic has focused on replacing the conventional controller, hence, relying entirely on DRL [7], [9]. This strategy poses a risk to safety-critical systems because: (i) DRL focuses on achieving the goal without considering stability, which could lead to catastrophic scenarios, (ii) DRL assumes fully observable states, which is rarely the case, and (iii) DRL is a black-box approach, making it difficult to predict behavior in anomalous situations. To overcome these problems we follow prior work that uses DRL in combination with a conventional controller as a form of hybrid architecture called *Residual Policy Learning (RPL)* [10] to create a complementary strategy [8], [11]. The expert policy can also act as a safety net and a correction factor, particularly in cases of unobserved states where the DRL agent would fail to optimize, which is demonstrated in our experimental results. The resulting synchronous policy is the superposition of both control strategies as also presented in [12], [13].

A second challenge of applying the DRL strategy directly to safety-critical applications is that random exploration can cause catastrophic failure. To address this, a *Cycle of Learning (CoL)* [9] method is used to efficiently train RPL. CoL is an improved version of *Behavioral Cloning (BC)*, where a base policy is trained to imitate the expert policy (conventional controller or a human expert) before DRL exploration. This helps the agent learn efficiently with limited data/random exploration.

A third problem for applying DRL in a complex and continuous state-action space (such as a real-world process involving multi-input variables) is that it learns through the entire dataset instead of the relevant data which can be challenging to converge. Using an RPL framework, one way to reduce the challenge of learning in massive state space is to split the state space into those parts that the conventional controller can manage and those parts where it struggles (abnormal situations), using Specialized Reinforcement Learning Agent (SRLA) proposed by the authors in [14]. We adopt the SRLA approach and further extend it to continuous state-action spaces and in the context of CoL with the RPL framework. Furthermore, through our proposed framework, we employ the Input-Output Hidden Markov Model (IOHMM) to automatically identify abnormal states, allowing the DRL agent to focus on learning the optimal policy in those targeted states. This approach is analogous to the options methodology [15], where each option identifies a specialization criteria within the hierarchical structure of reinforcement learning. The IOHMM approach is specialized for scenarios where identifying and handling abnormal states is essential. In contrast, the options methodology is a more general framework for hierarchical reinforcement learning, which can be applied to a wide range of tasks but may not focus explicitly on abnormal state detection and may not be as specialized for rare events.

*This work is supported by the EU-funded CISC project (MSCA grant: 955901) and conducted within the SCCH framework of the COMET Program managed by FFG. Kelleher's work is also partly funded by the ADAPT Centre under the SFI Research Centres Program (grant: 13/RC/2106_P2).

[1]Software Competence Center Hagenberg, Data Science, Austria `ammar.abbas@scch.at`
[2]Technological University Dublin, Computer Science, Dublin, Ireland
[3]ADAPT Research Centre, Dublin, Ireland

To the best of our knowledge, the performance of RPL has not yet been tested for application in safety-critical systems, nor has it been integrated with either CoL or SRLA. Consequently, the main contribution of this paper is to validate the proposed combination and demonstrate its efficiency in training an adaptable hybrid controller. We validate this framework on the Tennessee Eastman Process (TEP) control problem [16], [17]. In the first set of experiments (Section V), the combination of RPL (DRL + conventional controller) with CoL is shown to demonstrate the autonomous synchronization between the controllers. In the second experiment (Section VI), we demonstrate the autonomous activation of the controllers through SRLA, and in the last experiment (Section VII) the performance of the overall proposed hierarchical framework (RPL + CoL + SRLA) is validated through a comparative study in a complex scenario.

In the remainder of this paper, Section II discusses prior work and the main contributions of this paper, Section III presents the methodological framework, Section IV discusses the design of experiments, Section V presents the results of autonomous synchronization of controllers, Section VI discusses the results of autonomous activation of the DRL agent, and Section VII evaluates the comparison of our approach to the state-of-the-art, baselines, and traditional controller.

## II. RELATED WORK AND CONTRIBUTIONS

A number of works [7], [18]–[20] have compared conventional controllers with data-driven and adaptive DRL controllers or decision-support systems, noting that while DRL can solve non-linear complex and stochastic control problems, it has limitations such as lack of optimality guarantees and data requirements. Furthermore, in process or safety-critical industries, it is not recommended to completely replace the conventional controller and rely on such a learned policy that could fail in cases of unobserved states. Instead, DRL and conventional controllers should be used in complementary ways as suggested by [8]. One proposed approach is a hybrid architecture where the DRL agent handles high-level, offline decision-making while the conventional controller handles low-level, online control decisions, hence, working independently. However, our proposed approach in this context involves both controllers working together synchronously.

Residual Policy Learning (RPL) is a technique proposed by [10] that involves learning a residual on top of an existing controller, which can be more efficient than learning a policy from scratch. It was originally used in the context of robotics with similar work done by [12], [21]. Previous work has used RPL in control problems [11], [22], but it requires retuning the algorithm when the system encounters disturbance, setpoint change, or unobserved states. In our approach, we show another perspective of RPL, where the conventional controller acts as a safety net and correction factor on top of DRL in unobserved abnormal conditions. Furthermore, the prior work used a fixed weight ratio for the hybrid mix of superposed policy, whereas in our framework the proportion of the hybrid mix is autonomously determined by the multi-controller system.

To improve the performance of DRL, reference [9] integrates Behavioral Cloning (BC) and DRL with the goal of enabling efficient policy learning with limited data. This approach requires fewer interactions with the safety-critical systems by initializing the DRL agent with a base policy and then improving the policy in an online setting. Furthermore, in our work presented in [14], we introduced the Specialized Reinforcement Learning Agent (SRLA) hierarchical framework to enhance the convergence of DRL in intractable problems by focusing the learning process on targeted state-action spaces. We adopt these strategies and use them to enhance the performance of RPL. In this work, we also generalize SRLA, extend its application to a continuous action environment, and propose a method for autonomously identifying the target state through the state value function.

To validate the proposed framework (RPL + CoL + SRLA), we use the Tennessee Eastman Process (TEP) [16], [17]. In the realm of controlling the TEP, previous studies (such as [23], [24]) have explored the application of DRL-based control methods. However, a prominent challenge identified by these authors was the difficulty in attaining performance levels comparable to traditional controllers. To showcase the effectiveness of our approach, we conducted a comprehensive comparative analysis. Our findings demonstrate that our approach surpasses state-of-the-art methodologies in terms of performance and effectiveness.

## III. METHODOLOGICAL FRAMEWORK

### A. Preliminaries

The framework is composed of the following modules, (i) deep reinforcement learning, (ii) residual policy learning, (iii) cycle of learning, and (iv) input-output hidden Markov model.

*a) **Deep Reinforcement Learning (DRL):*** The Deep Deterministic Policy Gradient (DDPG) [4] is a model-free, actor-critic architecture [25] for DRL in continuous action space. The actor determines the policy, and the critic evaluates the policy to help the actor learn a better policy. A modified version of DDPG called Twin Delayed Deep Deterministic Policy Gradient (TD3) [26] is used in this study. The Q value modeled by the critic represents an expected total one-step reward also known as return $R_1$, discounted by a factor $\gamma$ with the objective to maximize the cumulative sum of rewards (or minimize the cost) while optimizing the policy $\pi$ as shown in Equation (1), where $s_t$ is the current representation of the state as observed from the environment and $a_t^A$ is the DRL agent's action policy:

$$Q^\pi(s, a^A) = \mathbb{E}_{a_t^A \sim \pi}\left[\sum_{t=0}^{\infty} \gamma^t R_1\left(s_t, a_t^A\right)\right] \quad (1)$$

The loss function used to optimize the critic ($L_Q$) through its parameters $\theta_Q$ is expressed as Equation (2), where $\theta_\pi$ represents the parameters of the actor-network that defines the policy and the other notations are similar to that of Equation (1).

$$\mathcal{L}_Q(\theta_Q) = \frac{1}{2}\left(R_1\left(s_t, a_t^A\right) - Q\left(\mathbf{s_t}, a_t^A\left(\mathbf{s_t} \mid \theta_\pi\right) \mid \theta_Q\right)\right)^2 \quad (2)$$

Where $R_1$ can be represented in terms of the critic network as shown in Equation (3), with $r_t$ being the one-step instantaneous reward added to the discounted sum of future rewards, $s_{t+1}$ being the next observed state after taking the action according to the policy, and the other notations are similar to the ones explained earlier.

$$R_1 = r_t + \gamma Q\left(s_{t+1}, a_t^A\left(s_{t+1} \mid \theta_\pi\right) \mid \theta_Q\right) \quad (3)$$

The loss function used to optimize the actor ($L_A$) through its parameters $\theta_\pi$ is defined in Equation (4):

$$\mathcal{L}_A(\theta_\pi) = -Q\left(\mathbf{s_t}, a_t^A\left(\mathbf{s_t} \mid \theta_\pi\right) \mid \theta_Q\right) \quad (4)$$

*b) Residual Policy learning (RPL):* RPL presented by [10] is a methodology that is designed to improve imperfect and nondifferentiable policies by learning a residual factor on top of a policy. It can be represented as:

$$C(s, \theta_\pi) = a_t^E + a_t^A\left(s_t \mid \theta_\pi\right) \quad (5)$$

In Equation (5), $a_t^E$ is the expert policy (i.e. conventional controller or a human-generated policy), $a_t^A$ is the DRL agent's policy, $\theta_\pi$ represents the parameters of the network that learns the residual policy over the expert, and $C$ is the combined policy as represented by the superposition of the baseline/expert policy and the agent policy.

*c) Cycle of Learning (CoL):* The Cycle of Learning (CoL) approach builds on top of *Behavioral Cloning (BC)*. BC as proposed by [27] is a model, where the agent tries to imitate the demonstrations provided by the expert behavior. The loss function for BC ($L_{BC}$) is represented in Equation (6), where $a_t^E$ is the expert policy same as in Equation (5), $a_t^A$ is the DRL agent's behavior policy, $\theta_\pi$ represents the parameters of the network that clones a behavior policy.

$$\mathcal{L}_{BC}(\theta_\pi) = \frac{1}{2}\left(a_t^E - a_t^A\left(s_t \mid \theta_\pi\right)\right)^2 \quad (6)$$

CoL addresses the problem of distributional drift. This mismatch is caused by the difference in state distribution between the policy initialized by pretraining through BC and fine-tuning the policy during the online training [9]. CoL integrates BC, actor, and critic loss from Equation (6), Equation (4), and Equation (2), respectively, during the pretraining as well as during the training as shown in Equation (7) to minimize the distribution drift.

$$\mathcal{L}_{CoL}(\theta_Q, \theta_\pi) = \mathcal{L}_{BC}(\theta_\pi) + \mathcal{L}_A(\theta_\pi) + \mathcal{L}_Q(\theta_Q) \quad (7)$$

*d) Input-Output Hidden Markov Model (IOHMM):* In an IOHMM (as described by [28]) the objective is to converge to an expectation maximization that determines the most probable hidden state (or sequence of hidden states) given the sequence of inputs and observations (outputs). Such a probability of the hidden states at any given time ($\zeta_t$) is shown in Equation (8), given the input sequence $U(u_1, u_2, \ldots u_n)$, observed sequence of variables $Y(y_1, y_2, \ldots y_n)$ and trained model parameters $\lambda$ (initial state, transition, and emission probability matrices). Where $n$ represents the sequence at the current time, $i$ represents a hidden state index ranging from 1 to $N$, $S$ is the notation for the estimated hidden state, and $x_t$ as presented in Equation (9) outputs the most probable hidden state at any instance of time given the sequence of observations. This most probable state in our framework will determine the desired state of interest to train and activate the DRL agent within the RPL framework.

$$\zeta_t(i) = P(x_t = S_i \mid U, Y, \lambda) \quad (8)$$

$$x_t = \underset{1 \leq i \leq N}{\operatorname{argmax}}\left[\zeta_t(i)\right] \quad (9)$$

*B. CoL-Based Specialized Deep Residual Policy RL*

In this section, we set out the framework that synergistically combines the four components described in Section III-A.

*a) IOHMM Training, and Initialization:* During the offline training phase, the expert trajectory serves two purposes: (i) behavioral cloning of the expert policy through CoL, and (ii) training the IOHMM. During the online training phase; the cloned policy now acts as the target actor and the initialization of the critic takes place. We train the critic using the Temporal Difference (TD) error as presented in Equation (2). This helps the critic to pre-train on the base policy so that the overestimation of the poor policy is avoided and the policy network is stabilized. The critic is trained on the reward associated with the superposition of the action through the expert and the pre-trained policy to help the DRL agent learn the residual action and improve the base policy as shown in Figure 1.

*b) Hidden State Classification Through Value Function (V):* During initialization, another essential step is performed by pretraining the state-value function ($V$) from the expert trajectory as shown in Equation (10), with $r_t$ being the one-step instantaneous reward added to the cumulative sum of future rewards discounted by $\gamma$ and $s$ defining the state of the environment as perceived by the DRL agent. The pretrained $V$ is used to classify the hidden states estimated through the trained IOHMM into the ones associated with poor performance or abnormal behavior as defined by the lower expectation of the value function $V$, which we call "hidden state specialization".

$$\mathcal{L}_V(\theta_V) = \frac{1}{2}((r_t + \gamma V(s_{t+1} \mid \theta_V)) - V(\mathbf{s_t} \mid \theta_V))^2 \quad (10)$$

*c) Online Training and Evaluation*: During online training and inference, the observed state is fed into the IOHMM for hidden state evaluation. If the estimated hidden state is the one which is specified through the state value function ($V$), then the Deep Residual Policy Reinforcement Learning (DRPRL) is activated to be trained and used in the superposed policy as shown in Figure 1. The DRL agent learns to work synchronously with the expert controller (conventional controller or a human agent) during the specified target states (such as abnormal situations in the process industries). The modified loss function used for training CoL-SDRPRL is shown in Equation (11), for simplification, the subscript notation of time ($t$) has been dropped. The algorithm of the proposed framework is shown as pseudocode in Algorithm 1. The source code is available at the project repository[1].

$$\mathcal{L}_{CoL-SDRPRL}(\theta_Q, \theta_\pi) = \mathcal{L}^*_{BC}(\theta_\pi) + \mathcal{L}^*_A(\theta_\pi) + \mathcal{L}^*_Q(\theta_Q) \quad (11)$$

where;

$\mathcal{L}^*_{BC}(\theta_\pi) = \frac{1}{2}\left(a^A(s^* \mid \theta_\pi) - a^E\right)^2$
$\mathcal{L}^*_A(\theta_\pi) = -Q\left(s^*, \left(a^E + a^A(s^* \mid \theta_\pi)\right) \mid \theta_Q\right)$
$\mathcal{L}^*_Q(\theta_Q) = \frac{1}{2}\left(R - Q\left(s^*, \left(a^E + a^A(s^* \mid \theta_\pi)\right) \mid \theta_Q\right)\right)^2$
$s^* = $ desired state identified through IOHMM by $V$

In Figure 1, the left part reflects offline training of the methodology, whereas the right part illustrates the online training phase. CoL is indicated by the "blue divot" region and the initialization of the value function and critic is indicated as the "red dotted diamond grid" region. The specialization components are indicated with the "yellow outlined diamond grid" regions. The offline specialization phase involves the output of the pre-trained state-value function which is fed to the output of trained IOHMM hidden states as shown with the arrow. For the online part, specialization is used to prune out the relevant desired data. This data is then used to train and activate the RPL which is represented by the "green dotted" region.

## IV. DESIGN OF EXPERIMENTS

In this section, we describe the experimental setup, simulation environment used, and case studies.

### A. Experimental Setup: DRPRL Controller

Deep Reinforcement Learning (DRL) has several components and its formulation varies between different environments and requires careful consideration, which we define in this section for our case study in process control.

[1]Project repository: https://anonymous.4open.science/r/CoL-SDRPRL

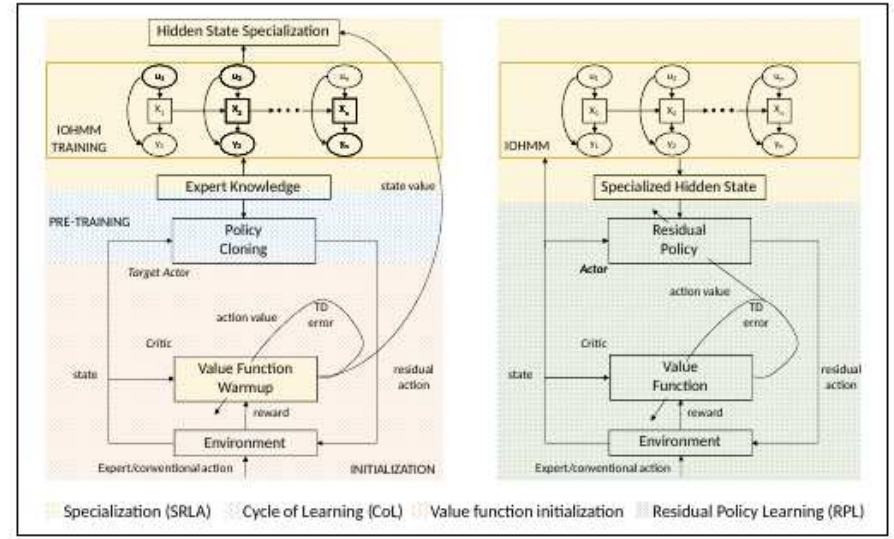

fig. 1: (left) Initialization of policy and critic and IOHMM training with classification through $V$. (right) Online training, activated on the specialized state.

*a) State*: In a Partially Observable Markov Decision Process (POMDP), the observed state alone may not be sufficient as the system's partial observability limits it [7]. Thus, the state observed by the DRL is different from the actual state of the environment. To address this, the study uses a tuple of the history of expert action concatenated with the history of process variables, which can have a length of up to $l$ as shown in Equation (12). The chosen history includes the current state ($t$) and only the previous trajectory ($t-1$), inspired by [10].

$$s_t := \langle (y_{t-l}, a^E_{t-l-1}), \ldots, (y_t, a^E_{t-1}) \rangle \quad (12)$$

*b) Reward*: The objective of an agent in a disturbance rejection problem is to find an optimal policy $\pi^*_\theta(s_t)$ that reduces tracking error and stabilizes the process with minimal deviation from the optimal set-point. This is achieved by integrating the goal into the DRL agent through a reward $r$ or cost $-r$ function, such as the negative $l_1$-norm of the set-point error [7]. Mathematically, for a system with $m_y$ process variables as inputs, it is shown in Equation (13).

$$r(s_t, a^A_t, s_{t+1}) = -\sum_{i=1}^{m_y} |y_{i,t} - y_{i,\text{sp}}| \quad (13)$$

*c) Layer Architecture*: The actor in the system uses two Long Short-Term Memory (LSTM) [29] layers with 64 and 32 neurons respectively, to incorporate POMDP state history and a fully connected neural network layer with a single sigmoid output unit. On the other hand, is composed of three fully connected layers with 64 ReLUs, 32 ReLUs, and a single linear unit in the output layer.

### B. Experimental Environment: Tennessee Eastman Process

We use the revised Simulink version of the Tennessee Eastman Process (TEP) plant [17] and modify it for our case studies. The process has 41 measurements (process variables) and 12 controllers (manipulated variables) (see tables 3-5 in [16]). The modified simulation environment is available in the project repository.

## Algorithm 1 CoL-SDRPRL Pseudocode

*Offline Training and Initialization*

**Input:** $B^E$: Expert trajectory ($S^E$), $N$: Hidden states
1: **for** pre-training steps = 1, ..., T **do**
2:   Compute $\mathcal{L}_{CoL}$ using Equation (7) and optimize
3: **end for**
4: **while** expectation-maximization tolerance $> 1 * 10^1$ **do**
5:   Compute expectation-maximization $\zeta$ (Equation (8))
6: **end while**
7: **for** warmup steps = 1, ..., W **do**
8:   Compute $L_V$ using Equation (10) and optimize
9: **end for**
**Output:** $\hat{x}$: Hidden states classification through $V$,
    $x^*$: Specialized hidden state

*Online Training and Inference*

**Input:** $B^E$: Expert trajectory ($S^E$), $B^A$: DRL buffer,
    Pre-trained $Q$ and $a^A$
10: **for** training steps = 1, ..., L **do**
11:   Compute $x_t$ using Equation (9).
12:   **if** $x_t$ is $x^*$ **then**
13:     $C(s, \theta_\pi) = a_t^E + a_t^A(s_t^* \mid \theta_\pi)$
14:     $Batch = 0.25 * B^E + 0.75 * B^A$
15:     Minimize $\mathcal{L}_{CoL-SDRPRL}$ using Equation (11)
16:   **else**
17:     Take expert's action ($a_t^E$)
18:   **end if**
19: **end for**
**Output:** $\pi_\theta^*(s_t^*)$: Optimal control policy (Equation (5))

*a) Single-Input Single-Output Setup:* In this setup, the DRPRL controller takes a single process variable, "flow feed A", as the input state and is affected by a step disturbance, "feed loss A". A single manipulated variable, "valve position feed A" is replaced by our proposed hybrid controller that is responsible for controlling the process variable and minimizing the disturbance. It further, demonstrates the possibility to integrate our framework within a set of different controllers.

*b) Multiple-Input Single-Output Setup:* In this setup, we take all the process variables (41) as the input state to the DRPRL controller. The disturbance and the manipulated variable replaced by the framework remain the same as in the previous setup.

### C. Case Studies

In the following case studies, the simulation runs with a step size of 0.01 simulation hour, during which the state information and reward are observed by the DRL agent for the policy. The control strategy is evaluated after every 5 episodes and after training at every step in between. The experiments are divided into three sets to address different research questions: (i) how the hybrid structure (RPL + CoL) works and complements each other, (ii) validating the convergence of an intractable continuous control problem using a specialized DRL agent, and (iii) comparing the proposed controller with baselines and examining the sensitivity of each element (RPL, CoL, and SRLA).

## V. EXP 1: AUTONOMOUS SYNCHRONIZATION OF CONTROLLERS

This set of experiments uses a SISO setup with the RPL + CoL framework without SRLA. The simulation runtime is 5 hours, and the disturbance's start and end times are randomized between the first and the second half of the simulation. Further, to verify the robustness and the complementary effects of the hybrid architecture, the disturbance magnitude is varied from 65% (over which the DRL[2] and DRPRL[3] agent is trained on) to 70% (unobserved state for the DRL and DRPRL agent). These specific percentages were chosen because these were the limits for the plant after which it observes emergency shutdown as the controller limits are reached.

The PID controller, DRL controller, and DRPRL controller responses are plotted in figures 2a, 2b, 3a, 3b, 4a, and 4b to disturbances of 65% and 70%, each. Additionally, the output of each controller (PID and DRL), extracted from the superposed signal of the DRPRL agent at each time step and their synchronization effects are demonstrated in figures 5a and 5b; for 65% and 70% disturbance, respectively.

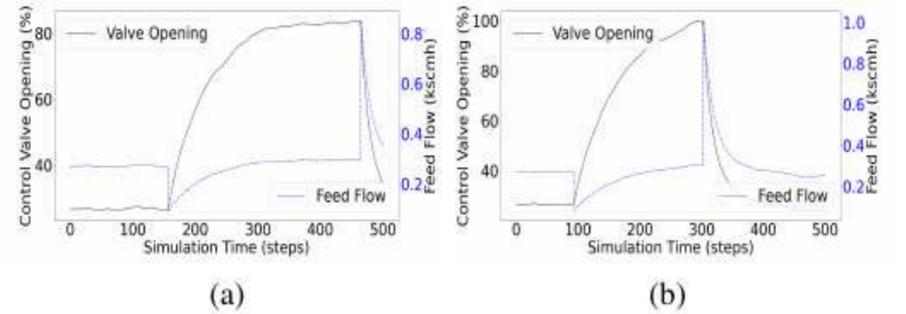

Fig. 2: PID response on (a) 65% and (b) 70% disturbance

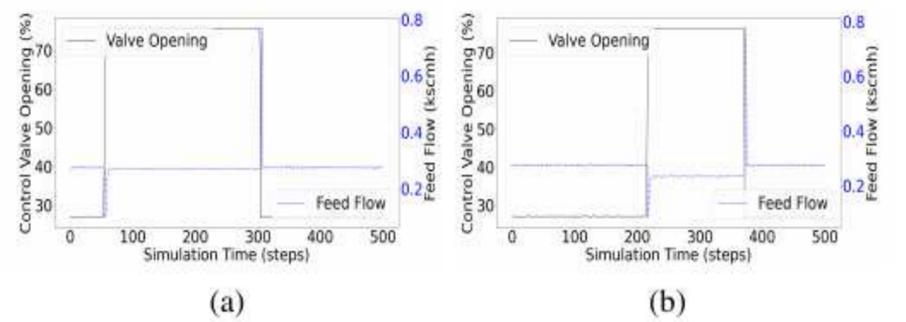

Fig. 3: DRL response on (a) 65% and (b) 70% disturbance

*a) Results and Discussion:* For the first set of experiments, the major criteria were (i) to observe the performance of the DRL controller that replaces the conventional controller in an unobserved scenario and (ii) to observe the synchronization between the conventional and the DRL controller in the proposed hybrid architecture and its performance in an unobserved scenario. The PID controller's disturbance rejection response has a delayed reaction time, risking the plant's emergency shutdown in the case of 70% disturbance (Figure 2b). In contrast, the DRL controller responds quickly and rejects the disturbance within a few time steps in the 65% disturbance case (Figure 3a). However, in the 70% disturbance case (Figure 3b), the DRL controller is unable to stabilize the system to its normal operating flow, and a residual control response is needed.

---
[2] DRL agent: actions that include only DRL.
[3] DRPRL agent: superposed DRL agent's action with the expert policy.

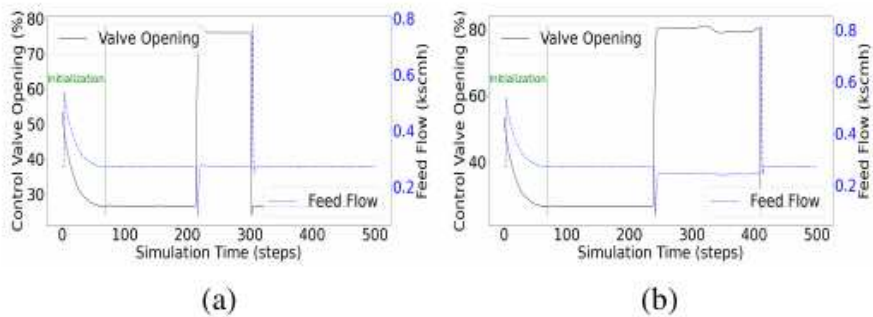

Fig. 4: DRPRL response on (a) 65% and (b) 70% disturbance

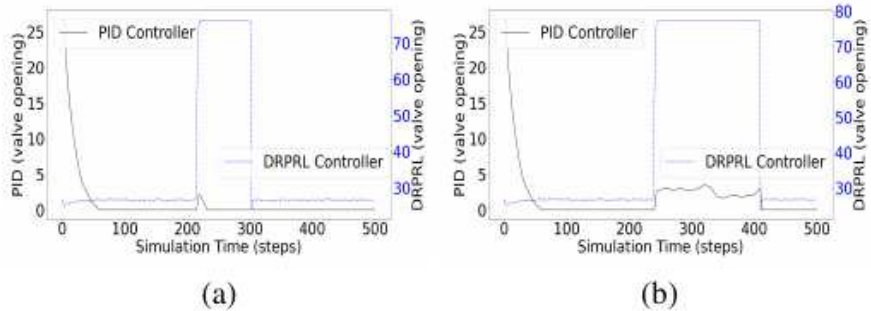

Fig. 5: Controller sync on (a) 65% and (b) 70% disturbance

In the hybrid architecture, we can observe an initialization[4] phase with the DRPRL controller, during which both the controllers synchronize as shown in Figure 4a. Once synchronized, the DRL controller takes over the PID and the PID response reaches zero as can be seen in Figure 5a. The advantage of having such an autonomous synchronization between the DRL agent and the PID controller is shown in Figure 4b, where the disturbance magnitude is increased. In this case, the PID controller comes back into effect and helps the DRL controller with the added correction to compensate for the remaining residual of the control as illustrated in Figure 5b.

To compare the results of the two baselines (PID and DRL controller alone) with the DRPRL controller, the simulation runs were evaluated for both cases of disturbances, and the average of episode returns was calculated for 10 runs after the 150 episodes of training steps. The evaluation was based on the episode returns after the initialization phase of the DRPRL case and the same was applied to the other cases having the same step-size delay for the accumulation of returns. The results are shown in Table II, where the DRPRL architecture outperforms the baseline controllers in a SISO process control setup.

---

[4]This initialization is based on the initial settings through which the controller initiates within the simulation script.

TABLE I: Comparison with the hybrid architecture

| Architecture | Average Episode Reward (10 runs) Disturbance Magnitude | |
|---|---|---|
| | 65% | 70% |
| PID | -17.71 | -23.54 |
| CoL | -2.58 | -12.25 |
| CoL+DRPRL | -1.85 | -6.25 |

## VI. EXP 2: AUTONOMOUS ACTIVATION OF CoL-SDRPRL AGENT

In this set of experiments, the MISO setup is utilized. The simulation runs for 2.5 hours in the simulation runtime and the disturbance starts randomly during the first half of the simulation and lasts till the end of the simulation. For these experiments, the convergence criteria are to be observed during normal and abnormal situations.

*a) CoL-Based DRPRL Controller:* The experiment runs for 250 episodes in this setup without having the specialization phase and the CoL-DRPRL is trained on the entire dataset. The controller responds to the normal and abnormal states and is unable to optimize in both situations.

*b) CoL-Based Specialized DRPRL Controller:* With the proposed framework (CoL-SDRPRL), the IOHMM is trained on the expert data (time series of process variables). The distribution of the hidden states defined by IOHMM is plotted on the process variable of "feed flow A" as shown in Figure A.1 of appendix A. It was observed that the IOHMM clearly identified the state that was associated with the abnormal situations (state 3), however, manual identification of such state is not practical in realistic scenarios, where the disturbance criteria are not known in advance and with a multivariate process it becomes difficult to isolate the variables that inherit disturbance. Therefore, the interpretation through the state value functions $(V)$ is used for classifying the abnormal state automatically, as defined in our framework. The abnormal situation is evaluated by the mean state value function with the negative score as shown in Table B.1 of appendix B. State 3 was identified as the state corresponding to the abnormal situations, and the DRPRL will now be specialized and activated, once such a state is predicted.

*c) Results and Discussion:* For the second set of experiments, the effect of autonomous switching between the DRL agent and the PID controller is observed and the comparison of the efficient training on the relevant data with the DRL agent trained on the entire data is validated. The response of the CoL-SDRPRL and autonomous activation of the agent during normal and abnormal scenarios is shown in Figure 6 after being trained on the relevant data for 50 episodes. The agent without specialization did not converge to the optimal policy and further deviated and performed poorly from the expert policy. However, in Figure 6, the PID controller works well in the normal state, and the CoL-SDRPRL agent is activated once an abnormal condition is observed. Furthermore, another important phenomenon can be observed as compared to the previous experimental results from Section V, the CoL-SDRPRL framework identifies the abnormal situation before acting on it and switches to the required controller. This prevents the disturbance from occurring and the plant operates at its normal behavior at all times, as opposed to observing the sudden spike and then recovering from it.

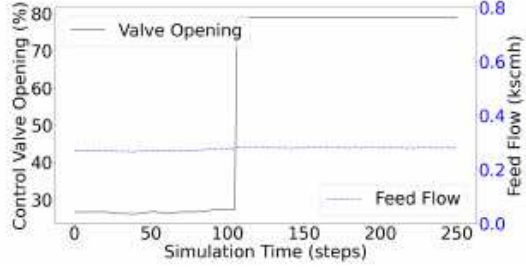

Fig. 6: CoL-SDRPRL trained on the relevant dataset (self-activated DRL agent)

## VII. EXP 3: COMPARATIVE EVALUATION

In this part of the experiment, the MISO setup is used and we compare the proposed architecture (CoL-SDRPRL) with baselines, where we use the possible combinations of the elements used in the methodology (BC, CoL, specialization, residual policy, and DRL) to identify the relevance of each factor. The disturbance level is set at 65%, and all models are trained with the same seed for 300 episodes. The initial experimental analysis and training curves are presented in Figure 7. The analysis shows that using DRL (specifically, TD3) or RPL framework without specialization or CoL results in inefficient training. Even with pre-training using BC, it does not help during the training due to the distribution shift. Although it eventually performs better than some baselines after a few episodes, this is not desirable in safety-critical applications. On the other hand, architectures with specialization outperform other baselines.

*a) Results and Discussion*: Based on the final evaluation; where we compare the proposed methodology with baseline combinations, the configurations that significantly outperformed the others were evaluated. An interesting factor to observe was that without residual policy, the policy started with a poor performance until, after a few episodes, it converged to its optimal policy. Table II evaluates the performance of each architecture and indicates its relevant features. The underlined abbreviation 'S' represents the architecture with specialization and can be observed that those configurations outperform the others, moreover, it is also validated that among them the ones that use RPL perform better, and lastly the CoL approach further increases the performance within them. Therefore, the effectiveness of each component is highlighted which enhances the overall control performance.

## VIII. CONCLUSION

This study aimed to investigate the performance of Residual Policy Learning (RPL) enhanced with a Cycle of Learning (CoL) and Specialized Reinforcement Learning Agent (SRLA). It integrates a conventional controller with a DRL controller for a safety-critical and intractable continuous control problem. Three sets of experiments were conducted: (i) to evaluate the synchronization between the controllers with their performance in an unobserved scenario, (ii) the convergence of the specialized DRL agent for the complex control problem, and (iii) the proposed controller's performance compared to baselines while examining the sensitivity of each element (RPL, CoL, and SRLA). The DRL controller responded quickly to disturbances, and the PID controller provided added corrections to compensate for the remaining residual of the control. The specialization approach allowed the DRL agent to focus on learning from and being applied to only the relevant (i.e., abnormal) data and proved to converge to an optimal policy unlike the one learned on the entire dataset. The overall results showed that the proposed framework (CoL-SDRPRL) outperformed the baselines and demonstrated robustness and complementary effects between the controllers, resulting in a stable system response.

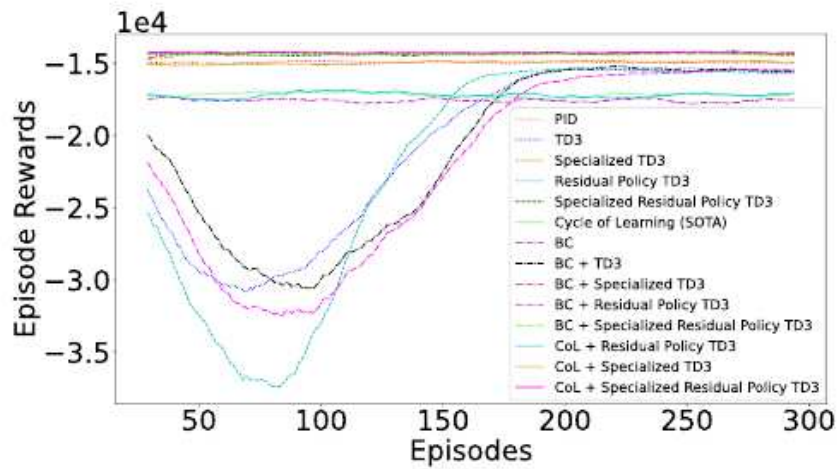

Fig. 7: Episode reward plot for comparative evaluation of CoL-SDRPRL with conventional methods

TABLE II: Comparison with baselines and state-of-the-art

| Method | Pre-Training Loss | Training Loss | Controller | Average Reward |
|---|---|---|---|---|
| CoL+SRPTD3 | $\mathcal{L}_{CoL-SDRPRL}$ | $\mathcal{L}_{CoL-SDRPRL}$ | $a^E + a^A$ | $-14274 \pm 350$ |
| BC+SRPTD3 | $\mathcal{L}_{BC}$ | $\mathcal{L}_Q + \mathcal{L}_A$ | $a^E + a^A$ | $-14340 \pm 324$ |
| SRPTD3 | None | $\mathcal{L}_Q + \mathcal{L}_A$ | $a^E + a^A$ | $-14342 \pm 334$ |
| STD3 | None | $\mathcal{L}_Q + \mathcal{L}_A$ | $a^A$ | $-14364 \pm 348$ |
| BC+STD3 | $\mathcal{L}_{BC}$ | $\mathcal{L}_Q + \mathcal{L}_A$ | $a^A$ | $-14370 \pm 360$ |
| PID | None | None | $a^E$ | $-14942 \pm 323$ |
| CoL+STD3 | $\mathcal{L}_{CoL}$ | $\mathcal{L}_{CoL}$ | $a^A$ | $-14956 \pm 294$ |
| CoL | $\mathcal{L}_{CoL}$ | $\mathcal{L}_{CoL}$ | $a^A$ | $-17178 \pm 701$ |
| CoL+RPTD3 | $\mathcal{L}_{CoL}$ | $\mathcal{L}_{CoL}$ | $a^E + a^A$ | $-17188 \pm 662$ |
| BC | $\mathcal{L}_{BC}$ | None | $a^A$ | $-17577 \pm 735$ |
| TD3 | None | $\mathcal{L}_Q + \mathcal{L}_A$ | $a^A$ | $-20702 \pm 6008$ |
| BC+TD3 | $\mathcal{L}_{BC}$ | $\mathcal{L}_Q + \mathcal{L}_A$ | $a^A$ | $-20716 \pm 5965$ |
| BC+RPTD3 | $\mathcal{L}_{BC}$ | $\mathcal{L}_Q + \mathcal{L}_A$ | $a^E + a^A$ | $-21745 \pm 6634$ |
| RPTD3 | None | $\mathcal{L}_Q + \mathcal{L}_A$ | $a^E + a^A$ | $-21845 \pm 8230$ |

# APPENDIX

## A. FIGURES

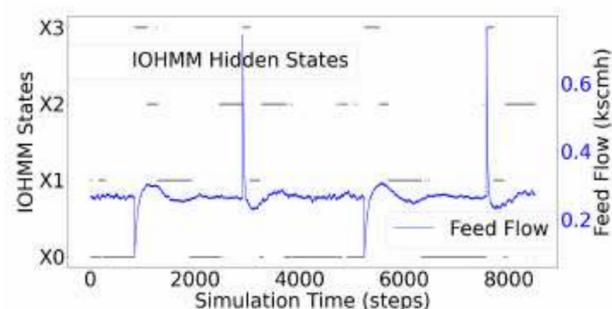

Fig. A.1: IOHMM state distribution

## B. TABLES

TABLE B.1: States classification based on the value function

| IOHMM states | mean state value | SD ($\sigma$) | 95% CI high | 95% CI low |
|---|---|---|---|---|
| $X_0$ | 7.58 | 2.30 | 7.62 | 7.53 |
| $X_1$ | 1.17 | 7.00 | 1.43 | 0.91 |
| $X_2$ | 5.39 | 3.92 | 5.51 | 5.27 |
| $X_3$ | -26.08 | 21.76 | -24.86 | -27.30 |